\documentclass[journal]{IEEEtran}
\usepackage{cite}
\usepackage{color,xcolor}
\usepackage{amsmath,amssymb,amsfonts}
\usepackage{algorithm}
\usepackage{algpseudocode}
\usepackage{graphicx}
\usepackage{textcomp}
\usepackage{wrapfig}
\usepackage{epstopdf}
\usepackage{textcomp}
\usepackage{array}
\usepackage{stfloats}
\usepackage{url}
\usepackage{booktabs}
\usepackage{threeparttable}
\usepackage{multirow}
\usepackage{multicol}
\usepackage{bm}
\usepackage{gensymb}
\usepackage{graphics} 
\usepackage{epsfig} 
\usepackage{times} 
\usepackage{balance}
\usepackage{stfloats}
\usepackage{setspace}

\begin{document}
		\title{Communication-Efficient Cooperative SLAMMOT via Determining the Number of Collaboration Vehicles}
		
	\author{Susu~Fang$^{2}$, and Hao~Li$^{*1,2}$ 
		\thanks{$^{1}$École d'Ingénieurs SJTU-ParisTech (SPEIT), Shanghai, 200240, China.}
		\thanks{$^{2}$Department of Automation, Shanghai Jiao Tong University (SJTU), Shanghai, 200240, China.}
		\thanks{$^{*}$Corresponding author: Hao Li (email: haoli@sjtu.edu.cn).}
	}

	
	\maketitle
\begin{abstract}
The SLAMMOT, i.e. simultaneous localization, mapping, and moving object (detection and) tracking, represents an emerging technology for autonomous vehicles in dynamic environments. Such single-vehicle systems still have inherent limitations, such as occlusion issues. Inspired by SLAMMOT and rapidly evolving cooperative technologies, it is natural to explore cooperative simultaneous localization, mapping, moving object (detection and) tracking (C-SLAMMOT) to enhance state estimation for ego-vehicles and moving objects. C-SLAMMOT could significantly upgrade the single-vehicle performance by utilizing and integrating the shared information through communication among the multiple vehicles. This inevitably leads to a fundamental trade-off between performance and communication cost, especially in a scalable manner as the number of collaboration vehicles increases. To address this challenge, we propose a LiDAR-based communication-efficient C-SLAMMOT (CE C-SLAMMOT) method by determining the number of collaboration vehicles. In CE C-SLAMMOT, we adopt descriptor-based methods for enhancing ego-vehicle pose estimation and spatial confidence map-based methods for cooperative object perception, allowing for the continuous and dynamic selection of the corresponding critical collaboration vehicles and interaction content. This approach avoids the waste of precious communication costs by preventing the sharing of information from certain collaborative vehicles that may contribute little or no performance gain, compared to the baseline method of exchanging raw observation information among all vehicles. Comparative experiments in various aspects have confirmed that the proposed method achieves a good trade-off between performance and communication costs, while also outperforms previous state-of-the-art methods in cooperative perception performance.
\end{abstract}

\begin{IEEEkeywords}
Communication-efficient C-SLAMMOT, cooperative intelligent vehicles, simultaneous localization and mapping (SLAM), moving object tracking (MOT)
\end{IEEEkeywords}

\IEEEpeerreviewmaketitle
\section{Introduction}
\IEEEPARstart{S}{imultaneous} localization, mapping, and moving object (detection and) tracking (SLAMMOT) coupling simultaneous localization and mapping (SLAM) and moving object tracking (MOT), allows for state estimation of ego-vehicle and moving objects simultaneously in dynamic environments \cite{wang2007simultaneous, bescos2021dynaslam, tian2023dl, ying2023imm}. SLAMMOT  can not only guarantee the performance of SLAM or MOT when the assumptions they rely on (the static world assumption for SLAM and the accurate ego-vehicle pose assumption for some MOT tasks) are difficult or even impossible to hold in complex dynamic environments, but also mutually benefit each other. However, single vehicle operation is still limited in many complex dynamic scenes, such as overtaking scenarios on bidirectional single-lane roads or ghost probe situations, which can easily lead to traffic accidents. For such complex scenes, multi-vehicle cooperative systems that take advantage of cooperative perception \cite{Li2011a, li2014multivehicle, Li2013b, fang2022lidar, wang2020v2vnet, xu2023cobevt} can overcome these limitations by interacting among multiple vehicles. The rapidly developing vehicular communication technologies and perception sensors can fairly support the research of cooperative technologies \cite{zeadally2020vehicular}. 

Inspired by the spirit of SLAMMOT and the spirit of cooperative perception, it is a natural idea to perform cooperative simultaneous localization, mapping, and moving object tracking (C-SLAMMOT) that combines the advantages of cooperative tasks and SLAMMOT to further augment the state estimation of the ego-vehicle and moving objects while overcoming the limitations of single-vehicle systems, thereby realizing coupled cooperative SLAM and multi-object tracking. Loosely-coupled methods perform the two estimation processes independently and ignore measurements from moving objects in the ego-vehicle state estimation, which do not benefit for each other \cite{stratidakis2019cooperative, miller2020cooperative}. This approach may not work well in dynamic environments where conventional cooperative SLAM is prone to degradation. In contrast, tightly-coupled methods extend cooperative SLAM to incorporate cooperative perception while integrating ego-vehicle state estimation, inter-vehicle relative pose, and object perception from multi-vehicle into a unified backend, achieving simultaneous estimation of the augmented states of the ego-vehicle and moving objects \cite{chang2011vision, chang2015exploiting, fang2024multi}.

However, how to optimize the trade-off between performance and communication cost is an inherent challenge in existing C-SLAMMOT solutions. Most methods transmit entire raw observations (images or point clouds) or a large volume of features among all collaborating vehicles. In practical applications, however, cooperative systems are often constrained by limited communication capacity, making it nearly impossible for them to handle such large data transfers in real-time. This can easily lead to network congestion and latency in the inter-vehicle communication. Moreover, as the number of collaboration vehicles increases, it becomes more challenging to handle in a scalable manner. Additionally, in some cooperative tasks, the information shared by certain cooperative vehicles may have little or no impact on the results, which can significantly waste valuable bandwidth \cite{xie2021rdc,hu2022where2comm}. Therefore, the goal should not solely focus on improving performance without evaluating the precious communication cost. Instead, it should aim to achieve a better balance between communication cost and performance. C-SLAMMOT performs localization, mapping, and object tracking simultaneously, which are fundamental functions for intelligent vehicles and are organically integrated in such applications. Many works \cite{chang2015exploiting,fang2024multi} primarily explore the mutual benefits between SLAM and MOT (as well as their corresponding cooperative tasks) in dynamic environments, and this beneficial relationship is mainly reflected in the state estimation of the ego-vehicle and objects. Also, these works can simultaneously generate maps, but they do not delve deeply into mapping. Thus, collaboration among vehicles in C-SLAMMOT primarily involves sharing local maps and poses to augment ego-vehicle pose estimation, and sharing moving object observations to enhance object perception. The exploration of communication efficiency will also take these two aspects into consideration.

In this paper, we present a communication-efficient C-SLAMMOT (CE C-SLAMMOT) method by determining the number of collaboration vehicles, which reduces communication costs while maintaining performance in enhancing the ego-vehicle's state estimation and cooperative object perception. The proposed method enables the ego-vehicle to initially share only global descriptors and spatial confidence maps to identify beneficial collaboration vehicles, then interact critical information with them, thereby reducing communication costs while achieving similar performance. To augment ego-vehicle state estimation in CE C-SLAMMOT, sequence global descriptor-based place recognition, which is less susceptible to dynamic scenes, is performed across multiple vehicles to identify the most similar matched vehicle. The corresponding local feature descriptors are extracted and shared to refine relative pose estimation, enhancing ego-pose in a communication-efficient manner. For cooperative object detection, the spatial confidence map, describing the likelihood of moving object presence, is used to determine collaboration vehicles with complementary areas, followed by sharing critical information with the ego-vehicle and fusing it via an axial attention module. This method not only reduces communication costs by dynamically adjusting the number of collaboration vehicles and interactive content, rather than transmitting raw observations among all vehicles as in existing C-SLAMMOT systems, but also maintains performance by focusing on critical information. Furthermore, the developed cooperative perception module outperforms the previous state-of-the-art (SOTA) network. The main contributions are as follows:

$\bullet$ We present a communication-efficient C-SLAMMOT method via reasonably determining the number of collaboration vehicles in both augmenting ego-vehicle state estimation and cooperative object perception, investigating the trade-off between performance and communication cost.

$\bullet$ We apply a descriptor-based method to determine the vehicle corresponding to the most similar point cloud frame for collaboration and estimate the relative pose, thereby reducing communication cost in the cooperative SLAM module.

$\bullet$ We develop a cooperative object detection module that adopts the spatial confidence map to select collaboration vehicles with complementary moving object information for the ego-vehicle's need, outperforming previous SOTA methods.
\section{Related Works}
\subsection{Coupled Cooperative SLAM and Object Tracking}
Single-vehicle systems have limitations in complex scenarios, such as crossroads with occlusions. Emerging cooperative technologies, including cooperative SLAM \cite{li2014multivehicle, fang2022lidar} and cooperative perception solutions \cite{wang2020v2vnet, xu2023cobevt}, address these limitations, inspiring the development of coupled cooperative SLAM and object tracking in either loosely or tightly coupled manners. In \cite{stratidakis2019cooperative}, a cooperative-aided localization tracking system was proposed to enable state estimation for multiple vehicles, including the ego-vehicle and tracked vehicles. Researchers then combined cooperative perception and localization technologies for cooperative driving systems, allowing autonomous vehicles to obtain real-time, accurate, and robust localization and perception information \cite{miller2020cooperative}. These loosely-coupled methods independently estimate the states of ego-motion and moving objects, with object detection results focused solely on tracking and not benefiting the ego-vehicle's state estimation.

In contrast, in \cite{chang2011vision}, researchers present a vision-based centralized cooperative simultaneous localization and tracking approach that integrates moving objects into the state vector to enhance EKF-based cooperative localization. Then, in \cite{chang2015exploiting}, researchers incorporate multiple hypothesis tracking into the centralized EKF-based multi-robot simultaneous localization and tracking (MR-SLAT) system to address challenging data association issues. In \cite{brambilla2022cooperative}, a joint technique for cooperative self-localization and multi-object tracking is proposed using a holistic and centralized approach, where graph theory describes the statistical relationships among agent states, object states, and observations. In \cite{fang2024multi}, a decentralized tightly-coupled cooperative simultaneous LiDAR SLAM and object tracking (C-SLAMMOT) system is introduced with a unified factor graph architecture that integrates information from the ego-vehicle, neighboring vehicles, and perceived object states to enhance state estimates for both the ego-vehicle and moving objects. Although these existing C-SLAMMOT methods perform well, they communicate entire raw observations among all collaboration vehicles, overlooking the trade-off between performance and communication cost.
\subsection{Communication-Efficient Cooperative SLAM}
Cooperative SLAM is a key component of C-SLAMMOT, utilizing communication technologies to share pose and static local map information for inter-vehicle relative pose estimation, thereby enhancing ego-vehicle state estimation. Many studies address this issue \cite{de2017survey}, with ranging sensor-based methods being a popular approach. These methods match two metric local maps from different vehicles to achieve indirect inter-vehicle relative pose estimation \cite{li2014multivehicle}. In \cite{li2020collaborative}, researchers use vision sensors for inter-vehicle relative pose initialization, while LiDAR sensors support continuous relative localization. In \cite{fang2022lidar}, the indirect inter-vehicle relative pose is estimated using only LiDAR point cloud matching. However, these applications require transmitting entire raw observations for matching, neglecting communication costs. Several cooperative SLAM applications focusing on communication issues have been proposed. In \cite{luft2018recursive}, a recursive multi-robot localization system with asynchronous pairwise communication is introduced, though it necessitates a complex data flow network for interaction. Additionally, researchers in \cite{xie2021rdc} present a real-time cooperative SLAM system aimed at improving computation and data transmission efficiency. This system separates inter-vehicle data association into place recognition, identifying common areas from different vehicles, and pose estimation, which matches local maps to estimate inter-vehicle relative poses. However, the use of DELIGHT \cite{cop2018delight} for point cloud segmentation-based place recognition in this work is unreliable in complex environments.

For LiDAR-based place recognition, traditional methods often rely on point cloud segmentation and segment descriptors \cite{dube2020segmap}, but these struggle in complex environments. Global descriptors aggregate local point cloud features but can be inaccurate with sparse data. Handcrafted descriptors like FPFH \cite{rusu2009fast} are limited under sparse conditions. Recent methods use deep learning to extract 3D point features, such as the DELIGHT descriptor utilizing LiDAR intensities \cite{cop2018delight} and PointNetVLAD \cite{uy2018pointnetvlad}, based on NetVLAD \cite{arandjelovic2016netvlad}. Additionally, converting 3D point clouds to 2D images can provide a more efficient way to extract features. This can be achieved through methods like projecting point clouds to planar coordinates and storing maximum height \cite{kim2018scan} or intensity \cite{wang2020intensity}. In the C-SLAMMOT task for dynamic environments, global descriptor extraction from single-frame methods is limited by dynamic interference. Sequence-based methods address this by representing sequences with descriptors, reducing matching costs and integrating temporal information to handle dynamic changes. SeqVLAD \cite{mereu2022learning}, based on NetVLAD, explores sequence descriptors and transformer integration but lacks interaction with temporal information. Recent work \cite{zhao2024learning} proposes a sequence descriptor that incorporates spatio-temporal information, using spatial attention within and across frames to capture feature persistence or change, making it more robust to moving objects and occlusions. After identifying common areas to build relationships between current point cloud scans and local maps from different vehicles, it can match similar local maps using extracted local features generated by methods such as SIFT \cite{lowe2004distinctive}, ORB \cite{rublee2011orb}, and SuperPoint \cite{detone2018superpoint}, solving relative pose estimation to augment ego-vehicle state.
\subsection{Communication-Efficient Cooperative Object Detection}
Cooperative object perception technology addresses the limitations of single-vehicle perception in complex scenes, but the trade-off between object perception performance and communication cost remains a challenge. Several solutions have been proposed to tackle this problem. In \cite{liu2020who2com}, researchers introduce a multi-stage handshake communication mechanism where a target agent with limited sensor data sends a compressed request, receives matching scores from other agents, and decides from whom to obtain information. Building on this, the When2com framework in \cite{liu2020when2com} optimizes communication timing to form a sparse communication graph. In \cite{wang2020v2vnet}, V2VNet combines end-to-end learning-based source coding with intermediate feature circulation from 3D backbones, utilizing a spatial-aware graph neural network for multi-agent feature aggregation. DiscoNet \cite{li2021learning} explores knowledge distillation, using 1D convolution to compress messages and aligning intermediate feature maps with an early-fusion teacher model. However, these works assume that once two agents collaborate, they must share perceptual information for all spatial areas equally, potentially wasting bandwidth since many areas may contain irrelevant information for object detection. In \cite{hu2022where2comm}, Where2comm introduces an efficient communication system guided by spatial confidence maps to determine where to communicate. It features a spatial confidence generator, a communication module for message packing and graph construction, and a message fusion module that utilizes multi-head attention to enhance feature maps. Additionally, CoBEVT in \cite{xu2023cobevt} employs sparse transformers to investigate vehicle connections and achieves SOTA performance in recent cooperative perception methods, with its fused axial attention module capturing local and global interactions across views and agents to effectively merge multi-view and multi-agent perception data within a transformer architecture.
\section{The Proposed Approach}
\subsection{Overview of Communication-Efficient C-SLAMMOT} 
\begin{figure*}[h]
	\centering
	\includegraphics[scale=0.45]{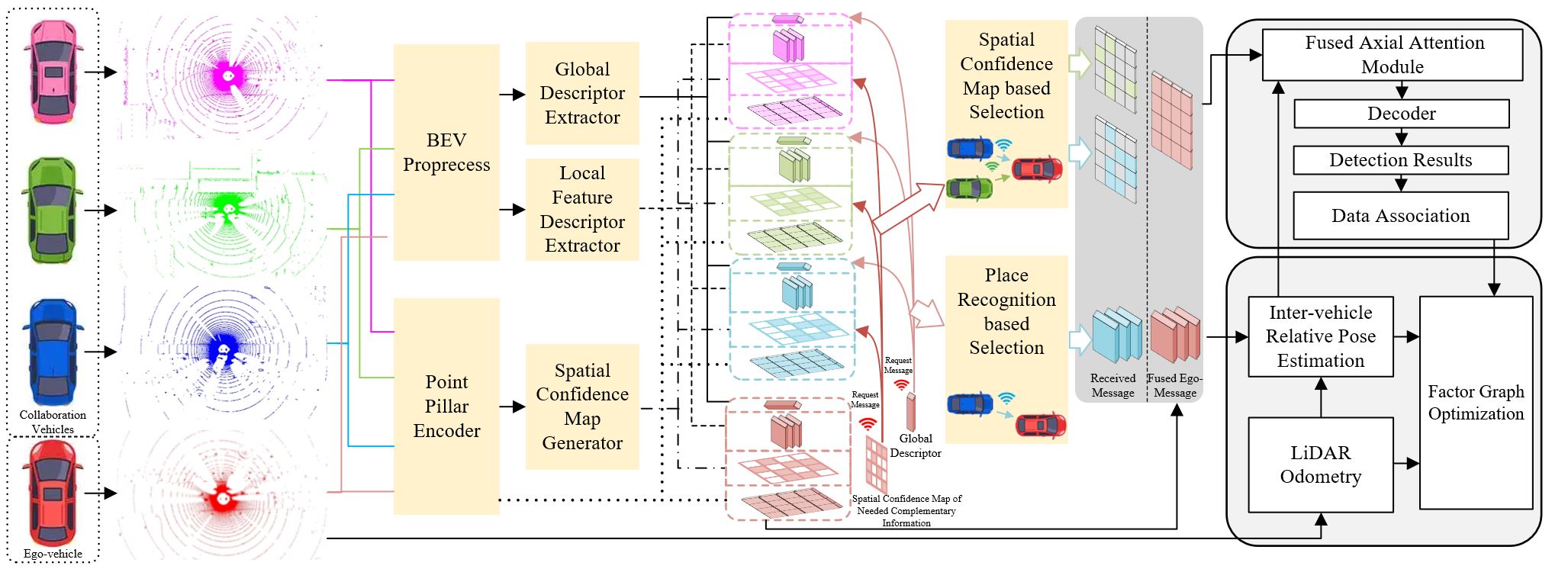}	
\caption{Architecture of the presented CE C-SLAMMOT.}
	\label{CSLAMMOT2}
\end{figure*}

The proposed CE C-SLAMMOT solution is an advanced version of our previous work in \cite{fang2024multi}. The architecture of the proposed solution is shown in Fig. \ref{CSLAMMOT2}, which mainly consists of a communication-efficient cooperative SLAM (CE C-SLAM) module, a superior and communication-efficient CoBEVT (CE CoBEVT) module that improves the existing SOTA cooperative object detection method, and a joint graph optimization backend. The 3D point cloud information from multiple vehicles serves as input for both the cooperative SLAM module and the cooperative object detection network. In the CE C-SLAM module, as detailed in Sec. III-B, the ego-vehicle sends the global descriptor generated in place recognition step as a request message to all neighbors to determine the collaboration vehicle with the most similar local maps, and then receives the corresponding extracted local feature descriptors from the selected neighboring vehicle to obtain inter-vehicle relative pose. In the CE CoBEVT module, the ego-vehicle sends the generated spatial confidence maps, which contain the needed complementary object information, to neighbors, enabling it to select collaborative vehicles with relevant information. After receiving the critical feature information from the selected vehicles, the ego-vehicle integrates them using a fused axial attention module for cooperative object detection, as detailed in Sec. III-C. Finally, both the state estimation from the cooperative SLAM module and the cooperative perception module are collectively optimized within a factor graph framework to jointly refine the state of the ego vehicle and objects. The two communication-efficient modules avoid wasting communication costs by sharing information with minimal impact or irrelevant content, continuously and dynamically selecting collaborative vehicles and interaction content while maintaining performance with a focus on critical interaction information. This achieves a good balance between performance and communication costs for related practical applications, whether in standalone cooperative SLAM, cooperative perception, or C-SLAMMOT tasks.
\subsection{Communication-Efficient Cooperative SLAM Module}
For the CE C-SLAM module in CE C-SLAMMOT, each vehicle performs its own LiDAR-based SLAM and interacts with neighboring vehicles in a decentralized manner to obtain inter-vehicle relative pose estimation, augmenting the ego-vehicle's state estimation. This module includes a single-vehicle LiDAR SLAM module, sequence global descriptor-based place recognition, and relative pose estimation, along with partial graph optimization. The sequence descriptor-based place recognition selects the optimal collaboration vehicle for determining with whom to establish inter-vehicle data association, while a local feature descriptor-based method obtains the corresponding relative pose estimation. The results are integrated into graph optimization as an inter-vehicle data association factor to jointly optimize ego-vehicle pose estimation. This module enhances ego-vehicle estimation by continuously and dynamically selecting the optimal vehicle, significantly reducing communication costs. At the same time, this selection method ensures the accuracy of relative poses among vehicles, maintaining the performance of the cooperative SLAM module and avoiding bandwidth waste from vehicles that contribute little but incur high communication costs during practical collaboration application. 
\subsubsection{Single Vehicle SLAM}
The single-vehicle SLAM can be any representative LiDAR SLAM framework. Given the multiple sub-modules in the proposed system, the lightweight nature of each module must be considered under conditions of limited computing resources. Therefore, LeGO-LOAM \cite{shan2018lego} is used for single-vehicle SLAM, as it maintains high localization and mapping accuracy while consuming fewer computing resources compared to LOAM \cite{zhang2017low}.
\subsubsection{Sequence Descriptor based Place Recognition}
In the CE C-SLAM module, the place recognition step identifies optimal data association matches between the ego-vehicle and neighboring vehicles. Traditional methods \cite{fang2022lidar, fang2024multi} typically require transmitting complete point clouds, which is communication-intensive, and some low-quality matches offer little improvement to ego-vehicle state estimation. To balance performance and communication costs, our approach minimizes communication by sending a sequence-based global descriptor that captures spatio-temporal features to all neighbors, focusing on the most similar frames in neighboring vehicles' local maps to ensure high-precision inter-vehicle relative pose estimation. The applied SeqVPR \cite{zhao2024learning} has also been shown to outperform recent SOTA methods in benchmark tests.

Before performing place recognition, we first generate bird's-eye view (BEV) images from point clouds. We use intensity projection images because LiDAR intensity readings reveal the reflectance structure of surrounding surfaces and vary for different objects \cite{kashani2015review, wang2020intensity}, enhancing place recognition by extracting information about moving and static objects in a dynamic environment. Specifically, LiDAR points are mapped to BEV grid indices within the LiDAR range, updating BEV intensity values and counts for each valid point. We then calculate and normalize the average intensity per grid cell for storage in the image pixels. In our application, the local maps of the ego-vehicle and neighboring vehicles serve as query and candidate data, respectively, for the SeqVPR method for descriptor extraction and place recognition. This sequence-based descriptor method uses a convolutional neural network to convert continuous raw frames into feature maps, which are divided into patches. A linear projection maps the patch features into embeddings. Single-frame embeddings are processed through a spatial transformer encoder with self-attention for spatial information, generating transformed embeddings. The embeddings from different frames within the sliding window are fused using a temporal transformer encoder to incorporate temporal information. Finally, the NetVLAD \cite{arandjelovic2016netvlad} layer aggregates these embeddings to generate a sequence descriptor. By sending global descriptors from the ego-vehicle's query frames as a request, the place recognition step identifies the most similar candidate frame from neighboring vehicles, determining the corresponding collaboration vehicle.
\subsubsection{Local Feature Descriptor based Relative Pose Estimation}
After determining the collaboration vehicle in place recognition, instead of transmitting the entire frame, local feature descriptors from the similar frame are sent to the ego-vehicle to estimate the inter-vehicle relative pose for enhancing the state of ego-vehicle. SuperPoint \cite{detone2018superpoint} is used to extract local features, utilizing convolutional layers to gather pixel data into an intermediate layer, followed by separate networks for feature detection and description. This network is chosen for two main reasons: its stable feature detection performance and its differentiable feature description model, making it suitable for training local feature description of generated BEV images. After receiving local features from the selected vehicle's candidate frame and extracting features from the ego-vehicle's query frame, SuperGlue \cite{sarlin2020superglue} is employed to match these features (including positions and descriptors). It uses a keypoint encoder to fuse contextual cues and applies alternating self- and cross-attention layers to obtain matching pairs. Since we lack a synthetic dataset for BEV images and the pre-trained SuperPoint network performs well, we use its preset weights. During training, the feature detection layers in SuperPoint are fixed, while its feature description layers and the entire SuperGlue are trained in a self-supervised manner.

After obtaining the image feature pairs with their corresponding position coordinates, we adopt the RANSAC mechanism to filter out incorrect feature matching pairs. Specifically, we solve the relative transformation relationship \begin{small}$\Delta T$\end{small} for each sampled coordinate pair set. This 2D relative transformation can be represented as a 3D transformation by zero-padding. According to the RANSAC score of \begin{small}$\Delta T$\end{small}, which counts the number of matches among all feature point pairs in the matching set under the transformation, we determine the credibility of the relative transformation. The transformation with the highest score is considered the final result. At the same time, the corresponding 3D pose of the determined neighbor vehicle's candidate frame in its map coordinate system \begin{small}${{\bf X}_{{{N},k}}}$\end{small} at the current moment $k$ is known. We can then obtain the determined neighbor-vehicle's pose \begin{small}${{\bf X}^{{M_{E}}}_{{N,k}}}$\end{small} in the ego-vehicle map coordinate system, and the relative pose of the neighbor-vehicle with respect to the ego-vehicle's map coordinate system \begin{small}${{\bf T}^{{E,k}}_{{N,k}}}$\end{small} can be further obtained as follows:
\begin{equation}
\footnotesize
\begin{aligned}
{{\bf X}^{{M_{E}}}_{{N,k}}}&=\Delta T \cdot  {{\bf X}^{{M_{E}}}_{{E,k}}}\\
{{\bf T}^{{E,k}}_{{N,k}}}&={{\bf X}^{{M_{E}}}_{{N,k}}}\cdot{{{\bf X}^{-1}_{{N,k}}}}
\end{aligned}
\end{equation}

And the determined neighboring vehicle continuously share its currently local features and its real-time pose data for collaboration, and ego-vehicle can solve the \begin{small}${{\bf T}^{{E,k}}_{{{N},k}}}$\end{small} to estimate its indirect pose \begin{small}${{\bf X}_{_{EN,k}}}$\end{small} for augmenting state estimation:
\begin{equation}
\footnotesize
\begin{aligned}
{{\bf X}^{{M_{E}}}_{{N,k}}}&= {{\bf T}^{{E,k}}_{{N,k}}}\cdot{{\bf X}_{{{N},k}}}\\
{{\bf X}_{_{EN,k}}}&=\Delta T ^{-1}\cdot{{\bf X}^{{M_{E}}}_{{N,k}}}
\end{aligned}
\end{equation}
\subsubsection{Graph Optimization in Cooperative SLAM}
The CE C-SLAM is based on factor graph optimization and uses  iSAM2 to integrate factors from the LiDAR odometry and inter-vehicle relative pose estimation to enhance ego-vehicle state. This module focuses on the performance of local cooperative SLAM tailored for congested or occluded intersections, without exploring loop closure detection in global optimization-based SLAM. Note the described graph optimization is only part of the optimization backend of the CE C-SLAMMOT, which covers more factors, as shown in Sec. III-D.
\subsection{Communication-Efficient Cooperative Object Detection}
The developed CE CoBEVT module addresses the limitations of single-vehicle perception in complex scenes, such as occlusions, while reducing communication costs and achieving superior performance. This network combines the strengths of Where2comm \cite{hu2022where2comm} and CoBEVT \cite{xu2023cobevt}, which are known for efficient communication and feature fusion in cooperative perception networks, respectively. CE CoBEVT uses a spatial confidence map as a request message to select collaboration vehicles and sends compact messages with complementary spatial features for the ego vehicle's needs. Messages are transmitted through a sparse communication graph, unlike methods that use full feature maps and fully connected graphs. The selected features from these vehicles are fused by an axial attention module to capture spatial interactions, improving detection accuracy. This continuous selection of vehicles and collaboration content avoids waste from sharing irrelevant information. Moreover, CE CoBEVT achieves SOTA results on the OPV2V dataset \cite{xu2022opv2v} and the real-world V2V4Real dataset \cite{xu2023v2v4real}, matching the results of previous SOTA methods while incurring lower communication costs. This is mainly due to its selection of fused data, reducing communication costs and minimizing potential false detections from irrelevant data.

Specifically, for the point cloud encoder in the cooperative object detection network, PointPillar \cite{lang2019pointpillars} extracts point cloud features \begin{small}${\bf F}_{i} \in \mathbb{R}^{H \times W \times C}$\end{small}. Inspired by \cite{hu2022where2comm}, a spatial confidence map \begin{small}${\bf{C}}_{i} \in [0,1]^{H \times W}$\end{small}, generated from a given feature map by a detection decoder, is applied. The needed complementary information request map \begin{small}${\bf{N}}_{i}= 1-{\bf{C}}_{i} \in \mathbb{R}^{H \times W}$\end{small} is negatively correlated with the spatial confidence map. First, the collaboration vehicles for perceiving moving objects are selected based on those that contain the most complementary object information for the ego-vehicle. A spatial confidence map generated by each vehicle's point cloud is converted into a dynamic mask map, marking the presence of objects in spatial areas as 1 and absence as 0 based on a score threshold. The complementarity of object information between vehicles is then calculated using the inclusion-exclusion principle, given by \begin{small}${V}_{e \to n}=f_{decide_1}(sum({\bf{D}}_{e}\cap {\bf{D}}_{n})/sum({\bf{D}}_{e}\cup {\bf{D}}_{n})),\in {\{0,1\}}$\end{small}, where \begin{small}${\bf{D}}_{e}$\end{small} and \begin{small}${\bf{D}}_{n}$\end{small} denote the dynamic mask maps converted from the corresponding spatial confidence maps \begin{small}${\bf{C}}_{e}$\end{small} and \begin{small}${\bf{C}}_{n}$\end{small} of the ego-vehicle and the neighbor vehicle, respectively. \begin{small}$\cap$\end{small} and \begin{small}$\cup$\end{small} denote the intersection and union operations. The determining function \begin{small}$f_{decide_1}$\end{small} calculates the complementary level \begin{small}${V}_{e \to n}$\end{small} with the ego-vehicle and selects vehicles whose object information is highly complementary to the ego-vehicle's perception. Next, the spatial confidence map is used to select the most complementary areas needed by the ego-vehicle, represented as \begin{small}${{\bf{M}}_{e \to n}}=f_{decide_2}({\bf{C}}_{n} \odot {\bf{N}}_{e}) \in \{0,1\}^{H \times W}$\end{small}. The function \begin{small}$f_{decide_2}$\end{small} selects the most critical areas based on the input matrix, with \begin{small}${\bf{M}}_{e \to n}$\end{small} representing the critical level at certain spatial locations. The selected feature map is then obtained as \begin{small}$\{{{\bf{M}}_{e \to n}} \odot {\bf{F}}_{i}\} \in \mathbb{R}^{H \times W \times C}$\end{small}, which provides spatially sparse yet perceptually critical information. Once the feature maps from collaboration vehicles are received, the features are warped by a differentiable spatial transformation operator to the coordinate system of the ego-vehicle and projected into a high-dimensional tensor \begin{small}$h \in \mathbb{R}^{N \times H \times W \times C}$\end{small}. These features are then fed into the fused axial attention module \cite{xu2023cobevt}, where the FuseBEVT encoder attentively fuses the received features. Finally, the fused features are used for classification and regression through two convolution layers for the final output.
\subsection{Joint Factor Graph Optimization}
\begin{figure}[h]
	\centering
	\includegraphics[scale=0.55]{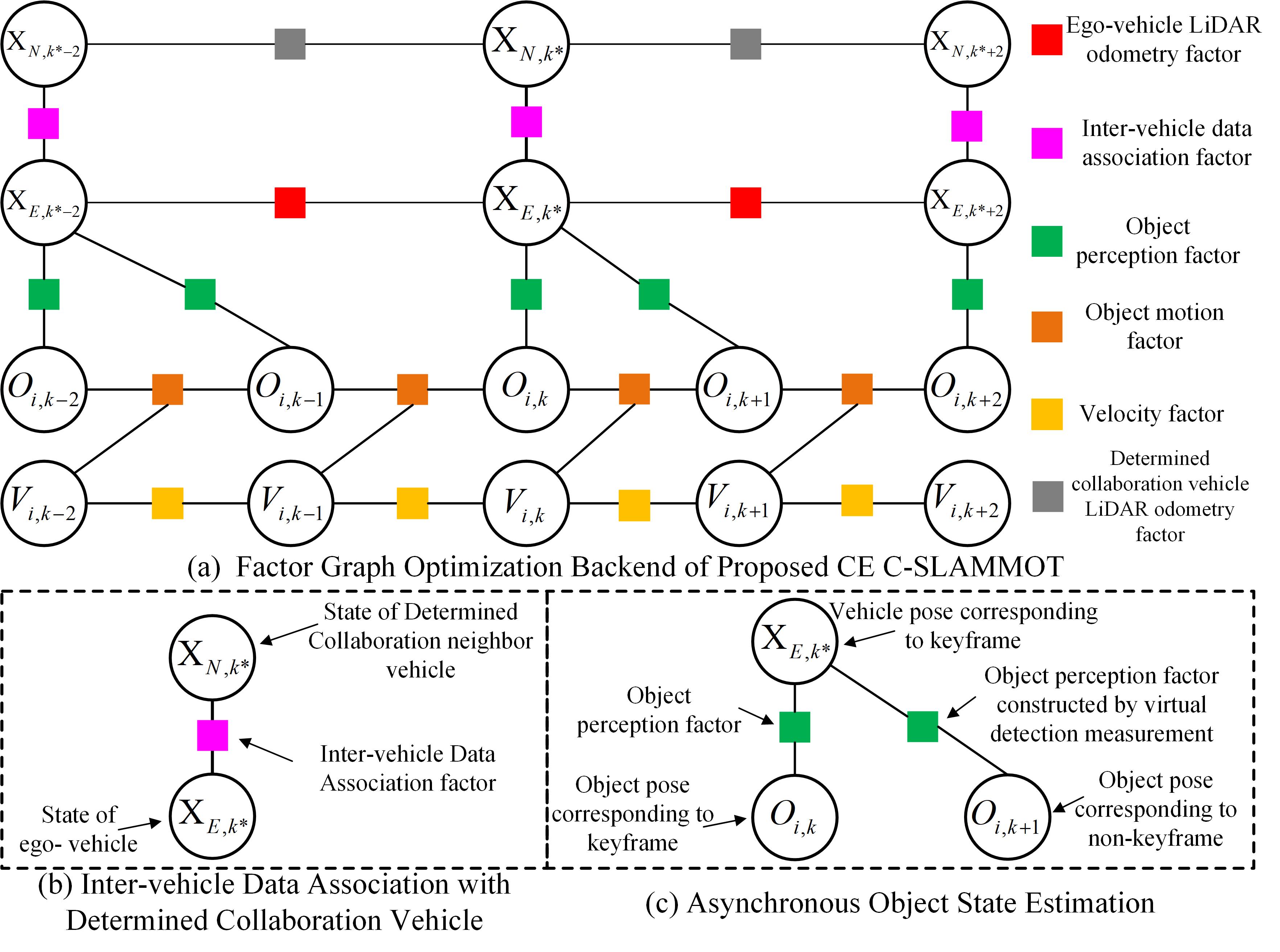}	
	\caption{The factor graph model in proposed CE C-SLAMMOT solution. (a) Joint factor graph optimization for state estimation of ego-vehicle and moving objects. (b) is a detail subfigure of inter-vehicle data association factor. (c) is an explanation subfigure of asynchronous object state estimation.}
	\label{FGO}
\end{figure}

The unified graph optimization backend couples cooperative SLAM with cooperative object detection to jointly optimize the state estimation of the ego-vehicle and moving objects, and achieves multi-object tracking. It extends the graph optimization backend of CE C-SLAM module. As depicted in Fig. \ref{FGO}(a), all variable nodes represent the estimated states. The ego-vehicle pose node \begin{small}${{\bf{X}}_{E,k^*}}$\end{small} at the timestamp of selected keyframes $k*$ is obtained through LiDAR odometry. The edge connecting two nodes represents the odometry constraint \begin{small}${{\bf{e}} _{odo}}$\end{small}:
\begin{equation}
\small
{{\bf{e}} _{odo}^{k^*-1,k^*}}=({{\bf{T}}_{k^*-1}^{k^*}} {{\bf{X}}_{k^*-1}}) ^{- 1}{{\bf{X}}_{k^*}}
\end{equation}

The edge between the ego-vehicle pose and determined collaboration vehicle pose \begin{small}${{\bf{X}}_{N,k^*}}$\end{small} is defined as the inter-vehicle data association constraint, where the indirect ego-vehicle pose estimation \begin{small}${{\bf{X}}_{_{EN,k^*}}}$\end{small} can be solved by the inter-vehicle relative pose estimation with the determined collaboration vehicle (see Sec. III-B for details). The factor can be expressed as following equation and is also illustrated in Fig. \ref{FGO}(b).
\begin{equation}
\small
{{\bf{e}} _{en}^{k^*}} = {{\bf{X}}_{_{EN,k^*}}^{{ - 1}}} {{\bf{X}}_{E,k^*}}
\end{equation}

The object perception factor connects the ego-vehicle pose node and the $i$-th moving object pose node \begin{small}${{\bf{O}}_{i,k}}$\end{small}. For efficiency in LiDAR odometry, keyframes are used for backend state estimation, meaning the vehicle state is updated only in keyframes, ignoring non-keyframe data, including moving object detections. Conventional methods typically estimate vehicle and object states synchronously, but factor graph optimization-based solutions do not have this limitation. Inspired by \cite{lin2023asynchronous}, we use asynchronous state estimation to utilize object information from non-keyframes. This transforms detections from non-keyframes into virtual measurements for the latest keyframes, based on the relative pose estimates obtained through scan matching between timestamps $k$ and $k^*$, as shown in Fig. \ref{FGO}(c). For data association in tracking, we use Gaussian mixture models (GMMs) for all detections and a max-mixture model to approximate the GMM sum, similar to the approach in \cite{poschmann2020factor,fang2024multi}. This implicit data association helps estimate object states without explicit assignment problem-solving, using detection results \begin{small}${{\bf{Z}}_{i,k}}$\end{small} obtained by cooperative perception with determined collaboration vehicles (see Sec. III-C for details). The object perception factor is denoted as:
\begin{equation}
\small
{{\bf{e}} _{op}^{k^*,k}} ({{\bf{X}}_{E,k^*}},{{\bf{O}}_{i,k}}|{{\bf{Z}}_{i,k}})=({{\bf{X}}^{{M_{E}}}_{E,k^*}})^{ - 1}{{\bf{O}}^{{M_{E}}}_{i,k}}({{\bf{Z}}_{i,k}})^{ - 1}
\end{equation}

Additionally, the edge between two object pose nodes is denoted as the object motion factor \begin{small}${{\bf{e}} _{mov,i}^{k-1,k}}$\end{small} shown in equation (6), where $f(\cdot)$ denotes the constant turn rate and linear velocity movement function of object states driven by its velocity \begin{small}${{\bf{V}}_{i,k}}$\end{small} in a time step \begin{small}$\Delta T$\end{small}. The edge constraints between two poses can be represented by velocity factors \begin{small}${{\bf{e}} _{v,i}^{k-1,k}}$\end{small} in equation (7), which assume a reasonable approximation that the object moves with a constant velocity over a short time.
\begin{equation}
\small
{{\bf{e}} _{mov,i}^{k-1,k}}=[f({{\bf{O}}_{i,k-1}},{{\bf{V}}_{i,k-1}},\Delta T)]^{ - 1}{{\bf{O}}_{i,k}}
\end{equation}
\begin{equation}
\small
{{\bf{e}} _{v,i}^{k-1,k}}={{\bf{V}}_{i,k-1}^{ - 1}}{{\bf{V}}_{i,k}}
\end{equation}

Finally, the joint factor graph optimization is a process that finds the best solution for the constructed factors based on the constraints in a factor graph. This process is defined as minimizing the sum of nonlinear least-square errors:
	\begin{equation}
\scriptsize
\begin{aligned}
{X^{*}}&=  \mathop {\arg \min }\limits_\chi \sum\limits_{k^*-1,k^*} {\left\| 	{{\bf{e}} _{odo}^{k^*-1,k^*}} \right\|_{{\sum _{odo}}}^2 }\!\!\!+\sum\limits_{k^*} {\left\| 	{{\bf{e}} _{en}^{k^*}} \right\|_{{\sum _{en}}}^2}\!\!\!+ {\left\| {{{\bf{e}} _p}({\chi _0})} \right\|^2}\\
  \!\!\!+&\sum\limits_{i,k-1\hfill\atop k^*,k\hfill}({\left\| {{\bf{e}} _{op}^{k^*,k}} \right\|_{{\sum _{op}}}^2}\!\!\!+{\left\| {{\bf{e}} _{mov,i}^{k-1,k}} \right\|_{{\sum _{mov}}}^2}\!\!\!+{\left\| 	{{\bf{e}} _{v,i}^{k-1,k}} \right\|_{{\sum _v}}^2})
\end{aligned}
\end{equation}

Where $\chi$ denotes the set of all variables, and \begin{small}${\left\| {{{\bf{e}} _p}({\chi _0})} \right\|^2}$\end{small} denotes the prior information error item. \begin{small}$\sum\nolimits_{{odo}}$, $\sum\nolimits_{{en}}$, $\sum\nolimits_{{op}}$, $\sum\nolimits_{{mov}}$\end{small} and \begin{small}$\sum\nolimits_{{v}}$\end{small} denotes the covariance or standard derivation matrix of each variable factor, respectively.
\section{Experimental Evaluation}
\subsection{Experiment Conditions}
In this section, we provide various comparative experiments with baseline methods to evaluate the performance of CE C-SLAMMOT in all respects. To assess the overall performance of the CE C-SLAMMOT, we conduct comparative experiments to compare and analyze different vehicle selection schemes to verify the rationality of our proposed vehicle selection scheme in different scenes, particularly in scenarios where multiple neighboring vehicles can collaborate. To highlight the advantages of the CE C-SLAMMOT over standalone communication-efficient cooperative SLAM or cooperative perception systems, we also perform comparative experiments on several important modules to evaluate our proposed solution for ego-vehicle and object state estimation under various constraints. We conducted experiments in various scenarios using the OPV2V dataset \cite{xu2022opv2v} and the real-world V2V4Real dataset \cite{xu2023v2v4real}, such as intersections and turning junctions.
\subsection{Communication-Efficient Cooperative SLAM Evaluation}
In the communication-efficient cooperative SLAM evaluation experiments, we first compare SeqVPR \cite{zhao2024learning} to existing SOTA place recognition methods in several dynamic scenes from OPV2V, including DELIGHT \cite{cop2018delight}, PointNetVLAD \cite{uy2018pointnetvlad}, NetVLAD \cite{arandjelovic2016netvlad}, and SeqVLAD \cite{mereu2022learning}. The area under the cumulative curve (AUC) score and the average recall at the top 1 and 3 retrieval candidates, based on the global descriptor distance of each pair, are shown in Table \ref{Tab.1}. If the centers of a pair are within a range of 10 m, they are considered positive. The results show that SeqVPR outperforms others due to its ability to extract more information and its lower susceptibility to dynamic environments. We also evaluated the performance of the applied local feature extraction and matching methods, as shown in Table \ref{Tab.2}. Comparing the precision and recall of feature point matching tasks in several scenarios reveals that the combination of SuperPoint and SuperGlue shows significant advantages over ORB and SIFT based methods, meeting the requirements for relative pose estimation.

Moreover, we compare ego-vehicle pose estimation among different methods: single-vehicle SLAM (LOAM \cite{zhang2017low}, LeGO-LOAM \cite{shan2018lego}), standalone cooperative SLAM \cite{fang2021adaptive} (without place recognition, using the entire point cloud for relative pose estimation), several CE C-SLAM methods (CE C-SLAM-1 to CE C-SLAM-5 applying the five aforementioned place recognition methods), the baseline C-SLAMMOT method \cite{fang2024multi}, and the proposed CE C-SLAMMOT solution. We use mean error (MEAN) and root mean square error (RMSE) as evaluation metrics. Experimental results from various scenes of the OPV2V and V2V4Real datasets are shown in Table \ref{Tab.3}. The table shows that several cooperative methods achieve smaller MEAN and RMSE compared to single-vehicle solutions. Additionally, some C-SLAMMOT-based solutions show performance improvements over methods that only conduct cooperative SLAM, demonstrating their superiority in dynamic environments. The comparison of CE C-SLAM, applying different place recognition methods, further shows that SeqVPR accurately identifies the most similar matching frames and corresponding vehicles, ensuring the accuracy of cooperative SLAM. The comparison among Cooperative SLAM, CE C-SLAM, baseline C-SLAMMOT, and CE C-SLAMMOT indicates that the presented method of reducing communication costs maintains good performance in ego-vehicle pose estimation. Furthermore, the CE C-SLAMMOT achieves superior accuracy in various scenes, indirectly reflecting the improvements of the CE CoBEVT module.
\begin{table}[h]
	\centering	
	\caption{Comparison of  Recall and AUC for place recognition methods}
	\label{Tab.1}
	\begin{center}
		\resizebox{8.7cm}{1.02cm}{
			\begin{tabular}{cccccccccc}
				\toprule
				\multirow{2}{*}{Methods} &\multicolumn{3}{c}{Test Scene-1}&\multicolumn{3}{c}{Test Scene-2}&\multicolumn{3}{c}{Test Scene-3}\\
				\cmidrule{2-4}\cmidrule{5-7}\cmidrule{8-10}
				&Recall@1&Recall@3&AUC&Recall@1&Recall@3&AUC&Recall@1&Recall@3&AUC\\
				\midrule
				DELIGHT\cite{cop2018delight}&0.829&0.845&0.852&0.739&0.748&0.871&0.365&0.509&0.533\\	
				\midrule					
				PointNetVLAD\cite{uy2018pointnetvlad} &0.955&0.966&0.925&0.793&0.807&0.883&0.607&0.674&0.715\\
				\midrule
				NetVLAD\cite{arandjelovic2016netvlad}  &0.927&0.938&0.872&0.821&0.828&0.970&0.604&0.723&0.845\\
    \midrule 
    SeqVLAD\cite{mereu2022learning} &0.968&\textbf{0.986}&0.888&0.982&\textbf{0.998}&0.967&\textbf{0.885}&0.916&0.885\\
    \midrule
    SeqVPR\cite{zhao2024learning} &\textbf{0.975}&0.982&\textbf{0.941}&\textbf{0.997}&\textbf{0.998}&\textbf{0.991}&\textbf{0.885}&\textbf{0.954}&\textbf{0.887}\\												
				\bottomrule
		\end{tabular}}
	\end{center}
\end{table}
\begin{table}[h]
	\centering	
	\caption{Comparison of Precision and Recall for different feature extraction and matching methods}
	\label{Tab.2}
	\begin{center}
		\resizebox{8.2cm}{0.82cm}{
			\begin{tabular}{cccccccc}
				\toprule
				Local Feature&Local Feature&\multicolumn{2}{c}{Test Scene-1}&\multicolumn{2}{c}{Test Scene-2}&\multicolumn{2}{c}{Test Scene-3}\\
				\cmidrule{3-4}\cmidrule{5-6}\cmidrule{7-8}
				 Extraction Methods& Matching Methods &Precision&Recall&Precision&Recall&Precision&Recall\\
				\midrule
				ORB\cite{rublee2011orb}&Brute force&0.374&0.325&0.403&0.379&0.343&0.308\\	
				\midrule					
				SIFT\cite{lowe2004distinctive} &Brute force&0.592&0.208&0.617&0.346&0.506&0.189\\
				\midrule
				SuperPoint\cite{detone2018superpoint}  &SuperGlue\cite{sarlin2020superglue}&\textbf{0.714}&\textbf{0.761}&\textbf{0.795}&\textbf{0.831}&\textbf{0.672}&\textbf{0.724}\\											
				\bottomrule
		\end{tabular}}
	\end{center}
\end{table}
\begin{table}[h]
	\centering
	\caption{MEAN/RMSE of ego-pose estimation by several methods}
	\label{Tab.3}
	\begin{center}
		\resizebox{8.7cm}{2.3cm}{
			\begin{tabular}{ccccc}
				\toprule
				\multirow{2}{*}{Methods} &\multicolumn{3}{c}{OPV2V	\cite{xu2022opv2v}}&V2V4Real\cite{xu2023v2v4real} \\
				\cmidrule{2-4}\cmidrule{5-5}
				&T-junction&Curved-road&Intersection&Turning-junction\\
				\midrule
				LOAM	\cite{zhang2017low}&0.289/0.292&0.401/0.436&0.456/0.474&1.582/1.591\\
				\midrule
				LeGO-LOAM \cite{shan2018lego}&0.272/0.281&0.388/0.421&0.432/0.457&1.614/1.623\\
				\midrule
				Cooperative SLAM \cite{fang2021adaptive}&0.238/0.250&0.299/0.315&0.381/0.398&1.179/1.158\\
				\midrule
				CE C-SLAM-1\cite{xie2021rdc}&0.264/0.267 & 0.357/0.379 & 0.427/0.443 & 1.433/1.458\\	
				\midrule
				CE C-SLAM-2\cite{uy2018pointnetvlad}&0.259/0.259 & 0.343/0.361& 0.404/0.434 & 1.392/1.412\\	
				\midrule
				CE C-SLAM-3\cite{arandjelovic2016netvlad}&0.252/0.256 & 0.331/0.353 & 0.393/0.429 & 1.344/1.313\\	
				\midrule
				CE C-SLAM-4\cite{mereu2022learning}&0.247/0.242& 0.314/0.327 & 0.394/0.403& 1.188/1.171\\					
				\midrule
				CE C-SLAM-5\cite{zhao2024learning}&0.242/0.235 &0.313/0.331&0.394/0.398&1.183/1.167\\				
				\midrule
				C-SLAMMOT \cite{fang2024multi}&\textbf{0.219}/\textbf{0.228}&0.217/\textbf{0.230}&0.343/0.360&0.878/0.904\\
				\midrule
				CE C-SLAMMOT (Ours)&0.226/0.231&\textbf{0.215}/0.239&\textbf{0.313}/\textbf{0.352}&\textbf{0.851}/\textbf{0.882}\\
				\bottomrule
		\end{tabular}}
	\end{center}
\end{table}

\begin{figure}[h]
	\centering
	\includegraphics[scale=0.60]{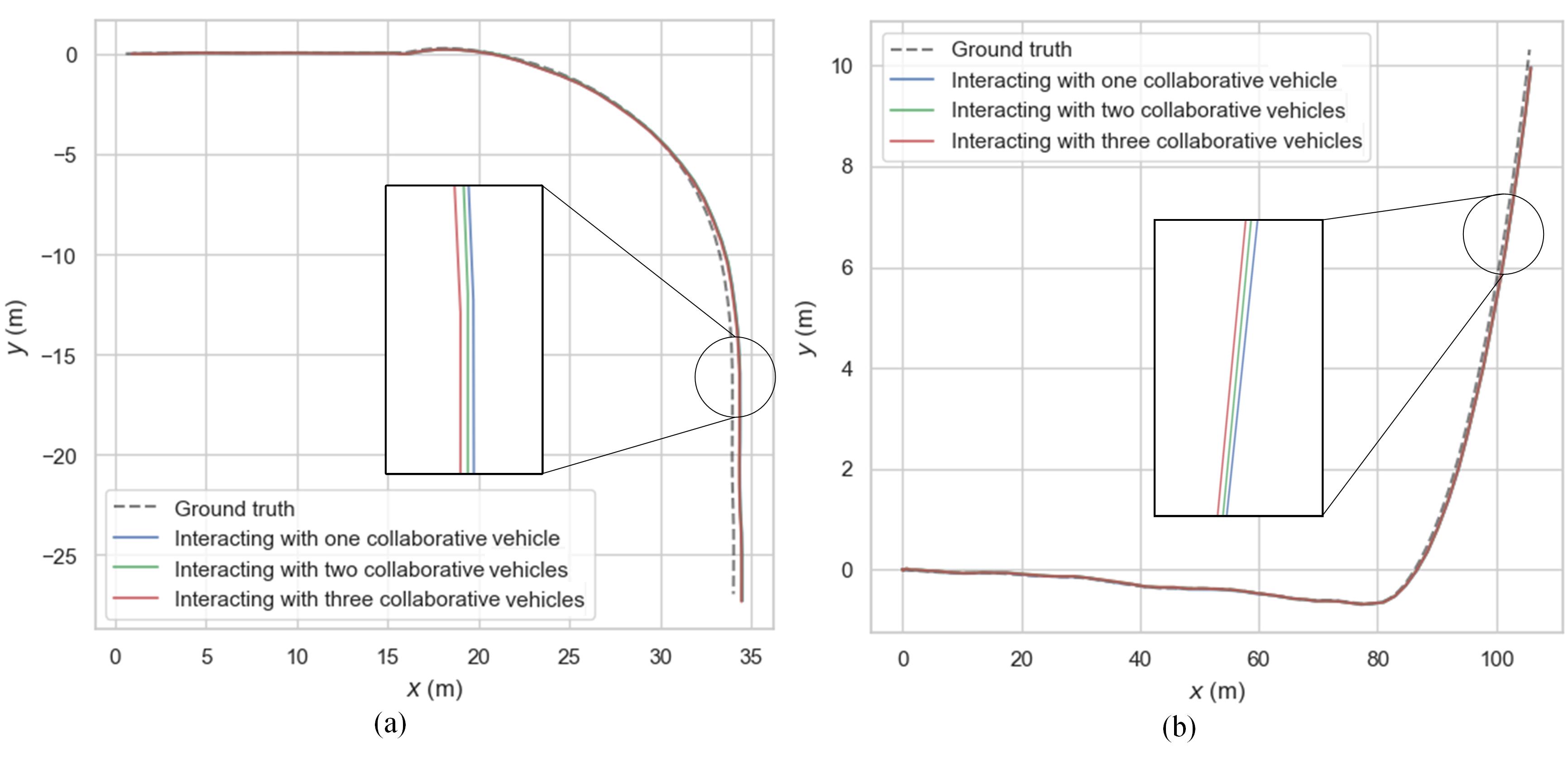}	
	\caption{Ego-trajectory error maps for presented method with different numbers of collaboration vehicles in cooperative SLAM module. (a) and (b) are the results in curved-road scene and intersection scene of OPV2V. The gray dashed line denotes the ground truth.}
	\label{Fig.3}
\end{figure}

In addition, Fig. \ref{Fig.3} shows the ego-vehicle trajectory error maps of the proposed method with different numbers of collaboration vehicles in two scenes. The vehicles are sorted by similarity obtained from place recognition. During the driving process, we dynamically select the vehicles with the highest similarity to achieve results with one collaboration vehicle. We then find corresponding point cloud frames of additional collaboration vehicles from the ranking to solve relative poses, obtaining results as more vehicles are added. It can be seen from the figure that continuous dynamic selection of optimal vehicles enhances ego-vehicle estimation, yielding effects similar to those achieved with multi-vehicle interactions. Other collaboration vehicles, having lower similarity, result in less accurate relative poses, contributing minimally to pose enhancement. However, this approach significantly reduces communication costs, as the number of interacting vehicles is proportional to the size of the transmitted messages. It is important to note that without reasonable vehicle selection, we cannot determine which vehicle has the greatest impact on pose enhancement, nor can we guarantee that randomly chosen vehicles will yield results comparable to those obtained through interactions with multiple vehicles. In curved-road and intersection scenes, the message size for interacting with one vehicle is about 286 KB and 250 KB, respectively (the entire point cloud size is 3.1 MB in OPV2V). Herein, we use local features from similar frames of the collaboration vehicle, averaging 276 keypoints in one scene and 241 in another, each with a descriptor dimension of 256.
\subsection{Communication-Efficient Cooperative Perception Module}
We conducted experiments on test datasets from OPV2V (Default Towns and Culver City) and the real-world dataset V2V4Real to evaluate the trade-off between detection performance and communication volume (CommVol) of the CE CoBEVT and several cooperative object detection methods. We measure message size in bytes on a log scale with base 2 to obtain the CommVol \cite{hu2022where2comm}. Detection performance is based on the Average Precision (AP) metric for vehicles, considering intersection over union (IoU) thresholds of 0.3, 0.5, and 0.7 (AP@0.3, AP@0.5, and AP@0.7), as shown in Table \ref{Tab.4}. Compared methods include When2com \cite{liu2020when2com}, V2VNet \cite{wang2020v2vnet}, Where2comm \cite{hu2022where2comm}, CoBEVT \cite{xu2023cobevt}, and CE CoBEVT. The results indicate that CE CoBEVT achieves a superior detection performance-communication cost trade-off across all communication volume choices and various cooperative perception tasks on all test datasets. Specifically, it shows substantial advancements over prior SOTA methods (CoBEVT) in V2V4Real and OPV2V, improving SOTA performance by 3.2\%/2.9\%/0.6\% (AP@0.3/0.5/0.7) on OPV2V Culver City and by 0.4\%/2.2\% (AP@0.5/0.7) on V2V4Real. This improvement is attributed to its selection of critical vehicles and spatial areas, reducing interference from irrelevant information for superior fusion. It also achieves the same detection performance (AP@0.5) as CoBEVT with significantly less communication volume: 32.6\% less on OPV2V Culver City, 6.5\% less on OPV2V Default Towns, and over 16.1\% less on V2V4Real.

\begin{figure}[h]
	\centering
	\includegraphics[scale=0.3]{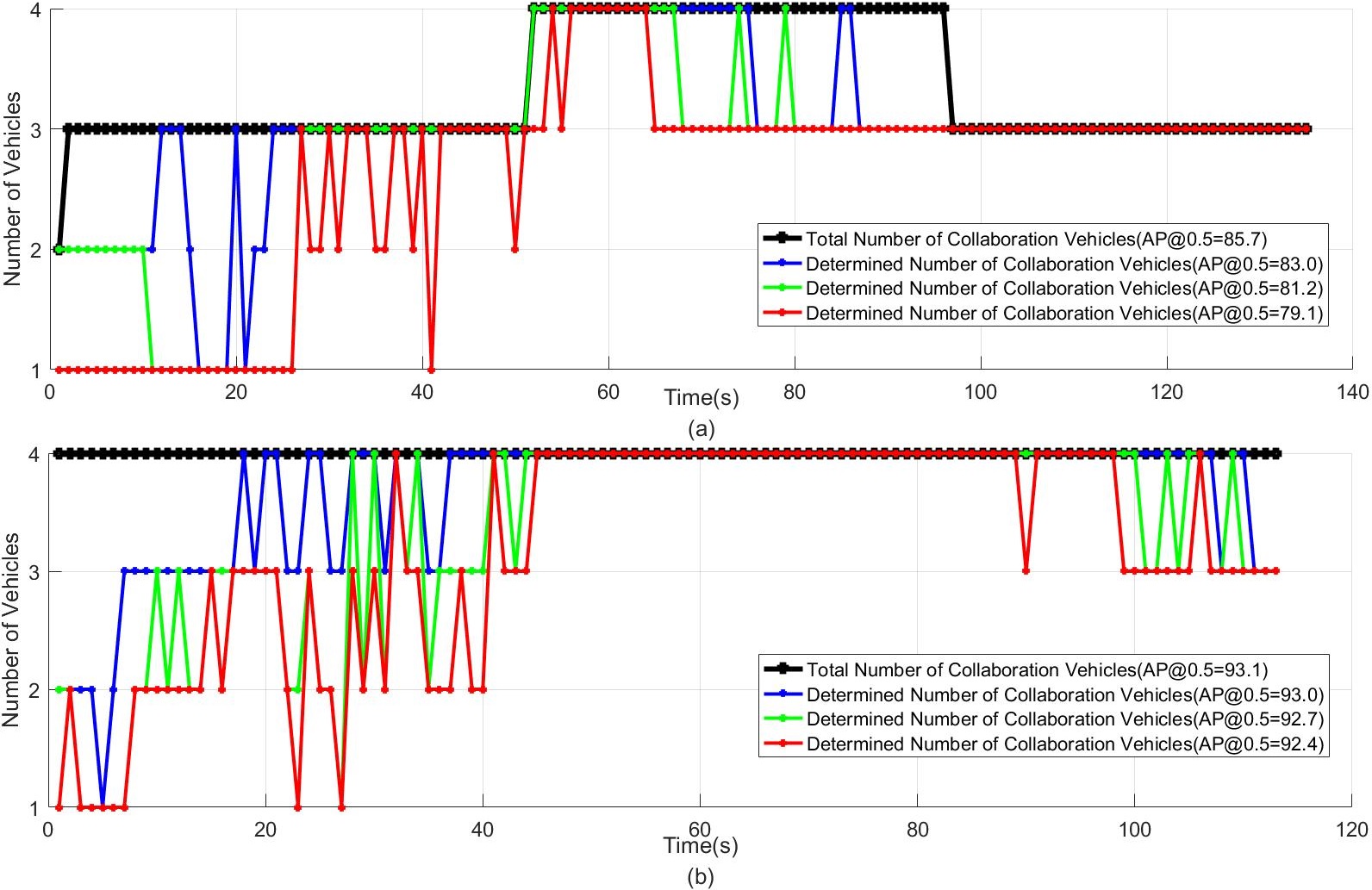}	
    \caption{The results of CE CoBEVT selecting cooperative vehicles. (a) and (b) denote the number of vehicles dynamically selected (including the ego-vehicle) for participation in cooperative perception under different detection precision (AP@0.5) in curved-road and intersection scenes, respectively. The black line denotes the total number of vehicles participating in cooperative perception without selection (not always four due to maximum communication distance limitations) with the corresponding detection precision. The blue, green, and red lines denote the changes in the number of cooperative vehicles with the corresponding detection precision under different selection situations.}
	\label{Fig.4}
\end{figure}

Table \ref{Tab.5} presents the detection performance (AP@0.5) of comparison methods in various test scenes with localization error. We use two methods to verify robustness to these errors. One method adds simulated localization noise (Gaussian noise with a mean of 0 and a standard deviation based on CE C-SLAM pose estimation; values for three scenes are 0.25, 0.35, and 0.85) to the ground truth pose during testing (Sim-Noisy AP). The other method uses real-time CE C-SLAM for pose estimation, replacing the cooperative perception module in CE C-SLAMMOT with the comparison model, followed by a statistical analysis of detection accuracy (Real-Noisy AP). Results indicate that CE CoBEVT outperforms other methods, even with localization errors. However, real-time pose estimation results are slightly worse than those with simulated noise, likely because Gaussian white noise does not account for some temporal-related noise in real pose estimation. Moreover, Table \ref{Tab.6} compares CE CoBEVT with the other methods in terms of the trade-off between perception performance (average precision of detection, AP@0.5), Multiple Object Tracking Accuracy (MOTA) and communication volume in different scenes. CE CoBEVT consistently achieves a superior performance-communication cost trade-off across two scenarios. For instance, in the intersection scene, CE CoBEVT uses only 34.8\%, 43.5\%, 63.9\%, and 80.8\% of the communication volume required by When2com, V2VNet, Where2comm, and CoBEVT, respectively, while achieving better detection precision. Additionally, Fig. \ref{Fig.4} shows the selection of collaboration vehicles by CE CoBEVT during the cooperative perception, highlighting the superior perception performance-communication cost trade-off.
\begin{table}[h]
	\centering
	\caption{Comparison of perception performance-communication cost trade-off on different datasets}
	\label{Tab.4}
	\begin{center}
		\resizebox{8.7cm}{2.6cm}{
			\begin{tabular}{cccccccc}
				\toprule
				\multirow{2}{*}{Methods} & \multicolumn{2}{c}{OPV2V Default Towns} &\multicolumn{2}{c}{OPV2V Culver City}&\multicolumn{2}{c}{V2V4Real}\\
				\cmidrule(r){2-3} \cmidrule(r){4-5}\cmidrule(r){6-7}
				&CommVol&AP@0.3/0.5/0.7&CommVol&AP@0.3/0.5/0.7&CommVol&AP@0.5/0.7\\
				\midrule
				When2com	\cite{liu2020when2com}&23.0&84.6/83.4/68.6&23.0&79.1/75.9/55.4&23.0&53.8/23.5\\	
				\midrule
				V2VNet \cite{wang2020v2vnet}&23.0&89.4/89.1/82.0&23.0&79.4/78.3/67.6&23.0&64.5/34.3\\	
			 \midrule
				CoBEVT \cite{xu2023cobevt}&23.0&90.7/90.2/83.5&23.0&83.2/81.8/72.2&23.0&66.1/34.4\\	
				\midrule
				\multirow{5}{*}{Where2comm} &12.6&85.8/83.9/72.9&13.8&80.2/76.5/61.9&13.2&63.1/34.1\\	
				\cmidrule(r){2-7}
				&13.8&86.9/84.1/73.9&14.9&83.3/77.6/62.9&14.6&63.2/34.3\\
								\cmidrule(r){2-7}
				&15.9&87.0/86.2/75.6&17.3&84.5/79.8/62.7&16.9&63.3/34.3\\
				\cmidrule(r){2-7}
				\cite{hu2022where2comm}&18.0&88.2/87.3/77.0&19.3&85.1/80.8/63.8&20.0&63.5/34.4\\
					\cmidrule(r){2-7}
				&20.5&89.2/88.3/77.9&21.2&85.6/81.9/65.8&21.7&64.1/34.8\\
					\cmidrule(r){2-7}
				&23.0&89.4/88.5/78.0&23.0&84.7/82.9/66.7&23.0&64.4/34.8\\			
				\midrule
					\multirow{5}{*}{CE CoBEVT}&12.2&86.7/85.5/75.4&13.8&82.4/79.9/67.3&12.3&63.4/35.4&\\
								\cmidrule(r){2-7}
				&15.6&87.5/86.3/76.8&14.7&83.7/81.5/69.6&15.5&64.3/35.8&\\
				\cmidrule(r){2-7}
				&17.1&88.6/87.6/78.9&18.2&84.6/82.5/69.2&16.7&65.5/36.1&\\
				\cmidrule(r){2-7}
				(Ours)&19.7&89.1/88.2/79.2&20.4&85.8/84.1/72.6&20.2&66.3/36.4&\\
				\cmidrule(r){2-7}
				&21.5&90.8/90.2/82.5&22.1&86.1/84.3/72.9&22.3&66.4/36.5&\\
				\cmidrule(r){2-7}
				&23.0&\textbf{91.6/91.1/83.7}&23.0&\textbf{86.4/84.7/73.3}&23.0&\textbf{66.5/36.6}\\	
				\bottomrule
		\end{tabular}}
	\end{center}
\end{table}
\begin{table}[h]
	\centering
	\caption{Evaluation result of cooperative object detection with localization error on several test scenes}
	\label{Tab.5}
	\begin{center}
		\resizebox{8.7cm}{1.02cm}{
			\begin{tabular}{ccccccccccc}
				\toprule
				\multirow{2}{*}{Methods} & \multicolumn{3}{c}{OPV2V Curved-road} &\multicolumn{3}{c}{OPV2V Intersection}&\multicolumn{3}{c}{V2V4Real Cross-intersection}\\
				\cmidrule(r){2-4} \cmidrule(r){5-7}\cmidrule(r){8-10}
				&Perfect&Sim-Noisy&Real-Noisy&Perfect&Sim-Noisy&Real-Noisy&Perfect&Sim-Noisy &Real-Noisy \\
				&AP@0.5&AP@0.5&AP@0.5&AP@0.5&AP@0.5&AP@0.5&AP@0.5&AP@0.5&AP@0.5\\
				\midrule
				When2com	\cite{liu2020when2com}&80.6&77.3&75.5&89.1&86.7&86.6&45.1&35.3&34.8\\	
				\midrule
				V2VNet \cite{wang2020v2vnet}&86.4&80.8&79.3&93.1&90.7&90.2&73.1&68.3&68.3\\
				\midrule
				Where2comm \cite{hu2022where2comm}&83.9&80.3&78.8&90.8&88.8&88.0&64.0&60.7&59.6\\
						\midrule
		CoBEVT \cite{xu2023cobevt}&90.2&86.8&84.8&96.0&92.9&92.5&72.6&68.1&67.5\\		
				\midrule
				CE CoBEVT (Ours)&\textbf{91.8}&\textbf{87.5}&\textbf{85.7}&\textbf{96.1}&\textbf{93.2}&\textbf{93.1}&\textbf{75.8}&\textbf{71.3}&\textbf{71.0}\\
				\bottomrule
		\end{tabular}}
	\end{center}
\end{table}
\begin{table}[h]
	\centering
	\caption{Comparison of perception performance-communication cost trade-off on several test scenes of OPV2V dataset}
	\label{Tab.6}
	\begin{center}
		\resizebox{8.0cm}{3.0cm}{
			\begin{tabular}{cccccccc}
				\toprule
				\multirow{2}{*}{Methods} &\multicolumn{3}{c}{Curved-road Scene} &\multicolumn{3}{c}{Intersection Scene}\\
				\cmidrule(r){2-4} \cmidrule(r){5-7}
				&CommVol&AP@0.5&MOTA&CommVol&AP@0.5&MOTA\\
				\midrule
				When2com	\cite{liu2020when2com}&23.0&75.5&72.7&23.0&86.6&84.4\\	
				\midrule
				V2VNet \cite{wang2020v2vnet}&23.0&79.3&77.1&23.0&90.2&88.6\\	
				\midrule
				CoBEVT \cite{xu2023cobevt}&23.0&84.8&82.6&23.0&92.5&91.7\\	
				\midrule
				\multirow{5}{*}{Where2comm} &13.9&75.6&72.9&11.6&86.9&84.7\\
				\cmidrule(r){2-7}
				&15.2&76.3&74.3&13.8&87.2&84.9\\
				\cmidrule(r){2-7}
				&18.0& 77.1&74.8&15.2&87.3&86.0\\
				\cmidrule(r){2-7}
				\cite{hu2022where2comm}&20.9&77.8&75.5&17.9&87.7&86.4\\
				\cmidrule(r){2-7}
				&22.1&78.7&75.9&20.8&87.9&86.9\\
				\cmidrule(r){2-7}
				&23.0&78.8&76.1&23.0&88.0&85.8\\			
				\midrule
				\multirow{5}{*}{CE CoBEVT}&14.4&77.6&73.8&11.1&88.6&86.5\\
				\cmidrule(r){2-7}
				&17.1&79.1&78.2&15.4&90.4&89.8\\
				\cmidrule(r){2-7}
						&20.1&83.0&81.0&18.3&92.4&89.9\\				
				\cmidrule(r){2-7}
		(Ours)&20.8&84.3&80.9&19.3&92.7&90.7\\
				\cmidrule(r){2-7}
				&22.2&85.3&82.7&21.1& 93.0&91.2\\
				\cmidrule(r){2-7}
				&23.0&\textbf{85.7}&\textbf{83.2}&23.0&\textbf{93.1}&\textbf{92.5}\\	
				\bottomrule
		\end{tabular}}
	\end{center}
\end{table}
\subsection{Overall Performance Evaluation}
\begin{figure}[h]
	\centering
	\includegraphics[scale=0.55]{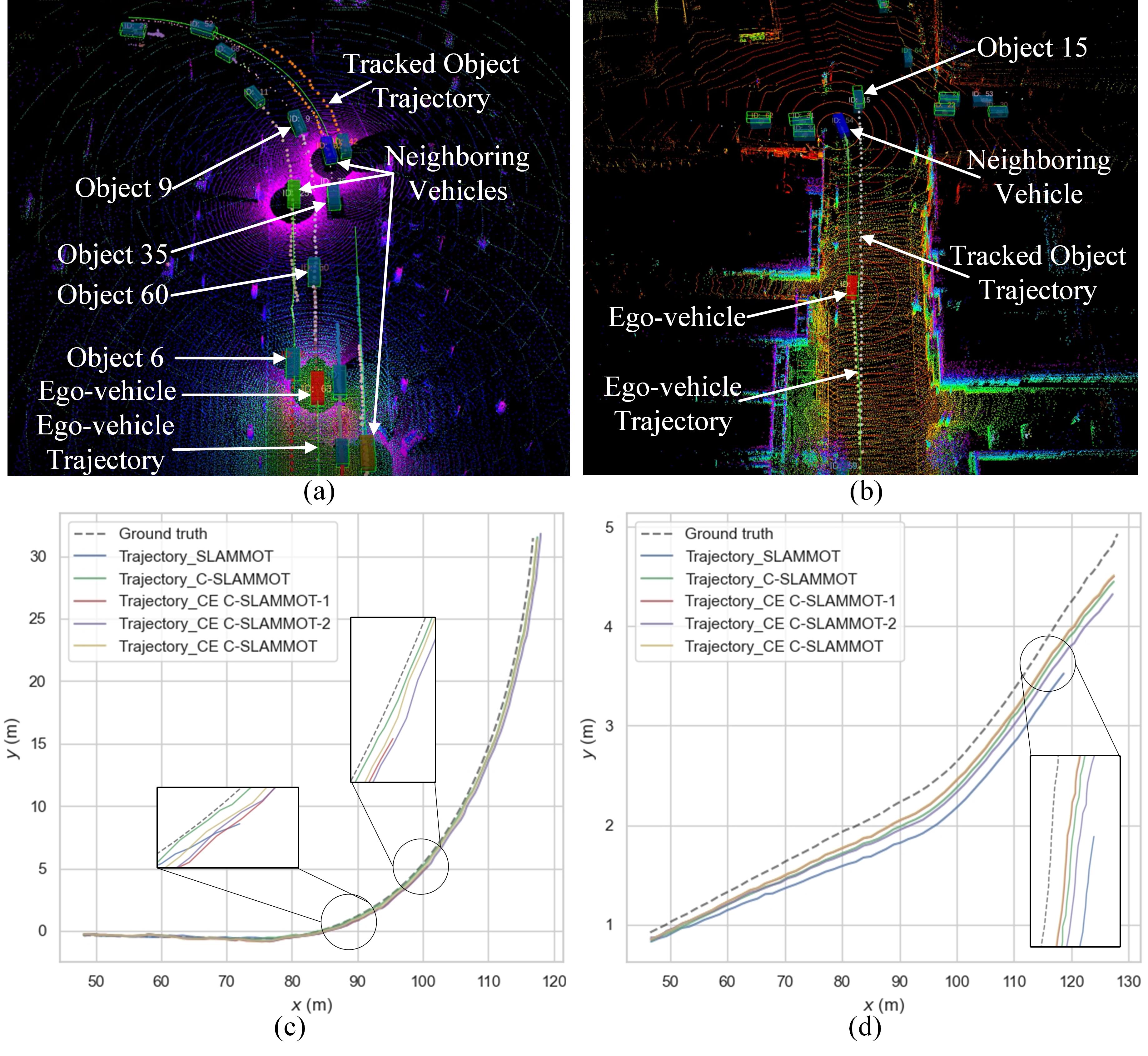}	
	\caption{The visualization results of the proposed CE C-SLAMMOT in the curved-road (a) of OPV2V and the cross-intersection (b) of V2V4Real. (c) and (d)  are the comparison of the estimated trajectories with different methods for object 9 in the curved-road and object	15 in the cross-intersection, respectively. The red, blue, green and olive rectangular solid denote the ego-vehicle and neighboring vehicles. The hollow green bounding box and midnight-blue rectangular solid denote detected and tracked objects with IDs. The green solid line denotes the pose estimation of the cooperative vehicles, while the dashed lines in different colors denote the tracking trajectories.}
		\label{Fig.5}
\end{figure}

We also assess the overall performance of the proposed method by evaluating the multi-object tracking accuracy (including RMSE for longitudinal and lateral positions, yaw angle, and velocity estimation) of several comparison solutions using different ways to select collaboration vehicles, as shown in Table \ref{Tab.7}. The compared methods include the single-vehicle SLAMMOT solution \cite{lin2023asynchronous}, the baseline C-SLAMMOT \cite{fang2024multi} (which communicates with all vehicles and transmits the raw observations, and uses CoBEVT in the cooperative perception module), the CE C-SLAMMOT-1 method \cite{liu2020who2com} that selects collaboration vehicles for cooperative SLAM and perception based solely on the place recognition step used in our proposed CE C-SLAM module, the CE C-SLAMMOT-2 method that determines collaboration vehicles based solely on spatial confidence maps used in our CE CoBEVT module \cite{hu2022where2comm,xu2023cobevt} (which presents results when the communication rate of the cooperative perception module is 0.808 for effectively selecting vehicles and comparing with C-SLAMMOT at the same detection accuracy). We analyze errors for three objects with different IDs: 6, 60, and 35, which represent short, medium, and long-distance vehicles, respectively. Table \ref{Tab.7} shows that the CE C-SLAMMOT performs similarly to the baseline C-SLAMMOT while reducing communication costs, as the baseline transmits the entire point cloud among all cooperative vehicles. CE C-SLAMMOT-1 selects only one vehicle for collaboration, which decreases perception performance and leads to ineffective detection and tracking of some occluded objects (such as object 9 in Fig. \ref{Fig.5}), thereby diminishing the auxiliary enhancement capability for ego-pose. CE C-SLAMMOT-2 also lowers communication cost compared to C-SLAMMOT, but the vehicles selected based on spatial confidence maps result in lower relative pose estimation accuracy, making its performance worse than the baseline. While CE C-SLAMMOT-2 saves communication cost at the expense of detection accuracy compared to CE C-SLAMMOT (with a communication rate of 1), it does not always select the optimal vehicle for cooperative SLAM, increasing the communication cost of the module without significantly enhancing the ego-vehicle's state estimation. The estimated trajectories of objects for different methods are shown in Fig. \ref{Fig.5}.
\begin{table}[h]
	\centering
	\caption{Evaluation of multi-object tracking for different methods in curved-road scene of OPV2V dataset}
	\label{Tab.7}
	\begin{center}
		\resizebox{8.0cm}{3.02cm}{
			\begin{tabular}{cccccc}
				\toprule
				\multirow{2}{*}{Methods}&\multirow{2}{*}{Object ID} &RMSE&RMSE&RMSE&RMSE\\
& &(long/m)&(lat/m)&(rad)&(m/s)\\
\midrule			
					\multirow{3}{*}{SLAMMOT}&6&0.447 &0.113 &0.157 &0.503\\
\cmidrule{2-6}
&35&0.514 &0.166 &0.201& 0.572\\		
\cmidrule{2-6}
\cite{lin2023asynchronous}&60	&0.467& 0.145& 0.190& 0.521\\					
\midrule				
				\multirow{3}{*}{C-SLAMMOT}&6&0.328&0.078&0.104&0.383\\
				\cmidrule{2-6}
				&35&\textbf{0.428}&0.115&\textbf{0.169}&\textbf{0.452}\\
								\cmidrule{2-6}
				\cite{fang2024multi}&60&0.362&0.108&0.142&\textbf{0.421}\\								
				\midrule				
			\multirow{1}{*}{CE C-SLAMMOT-1\cite{liu2020who2com}}&6&0.332&\textbf{0.075}&0.107&\textbf{0.373}\\
			\cmidrule{2-6}
		(Only place recognition&35&0.459&0.131&0.192&0.488\\	
						\cmidrule{2-6}
	based selection )&60&0.371&0.112&0.141&0.426\\
							\midrule											
				\multirow{1}{*}{CE C-SLAMMOT-2 \cite{hu2022where2comm}}&6&0.342&0.093&0.118&0.398\\
	\cmidrule{2-6}
	(Only sptical confidence map &35&0.452&0.120&0.181&0.471\\		
	\cmidrule{2-6}
	 based selection )&60&0.374&0.114&0.146&0.440\\					
	\midrule									
				\multirow{3}{*}{CE C-SLAMMOT}&6&\textbf{0.326}&\textbf{0.075}&\textbf{0.103}&\textbf{0.373}\\
				\cmidrule{2-6}
				&35&0.432&\textbf{0.111}&0.174&0.458\\	
								\cmidrule{2-6}
				(Ours)&60&\textbf{0.361}&\textbf{0.105}&\textbf{0.136}&0.426\\		
				\bottomrule
		\end{tabular}}
	\end{center}
\end{table}
\section{Conclusion}
This paper presents a communication-efficient cooperative SLAMMOT method by determining the number of collaboration vehicles. This solution determines the optimal collaboration vehicle through sequence place recognition-based selection in the cooperative SLAM module and selects critical collaboration vehicles through the spatial confidence map in the cooperative perception module. And ego-vehicle interacts with these vehicles to exchange only the critical information needed. This approach achieves similar or even better performance to baseline solutions while significantly reducing communication cost. Experimental results demonstrate the trade-off between performance and communication cost of the proposed method in various aspects.

In addition, this work presents methods to select suitable collaboration vehicles for ego-vehicle state estimation and cooperative object perception. In the future, the relationship between the number of vehicles selected for these two aspects can be further explored. And the investigation of appropriate weight settings to determine a unique number of collaboration vehicles when both aspects are considered together.
\ifCLASSOPTIONcaptionsoff
\newpage
\fi
\bibliographystyle{IEEEtran}
\bibliography{Refs_Index}

\newpage
\balance

\begin{IEEEbiography}[{\includegraphics[width=1in,height=1.25in,clip,keepaspectratio]{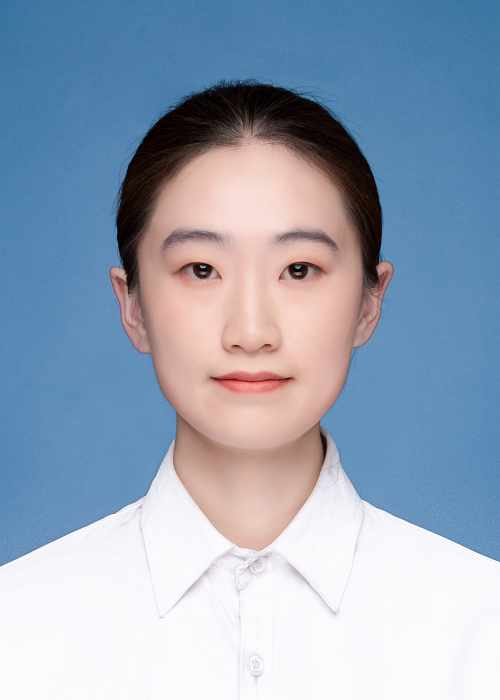}}]{Susu Fang},
currently a Ph.D. candidate at the Department of Automation, Shanghai Jiao Tong University, Shanghai, China. She obtained the M.Eng. degree and B.Eng. degree from Shandong University and Nanjing Agricultural University, China, in 2019 and 2016 respectively. Her research interests are in autonomous vehicle localization and perception, multi-sensor data fusion, and multi-vehicle cooperative intelligent systems.
\end{IEEEbiography}

\begin{IEEEbiography}[{\includegraphics[width=1in,height=1.25in,clip,keepaspectratio]{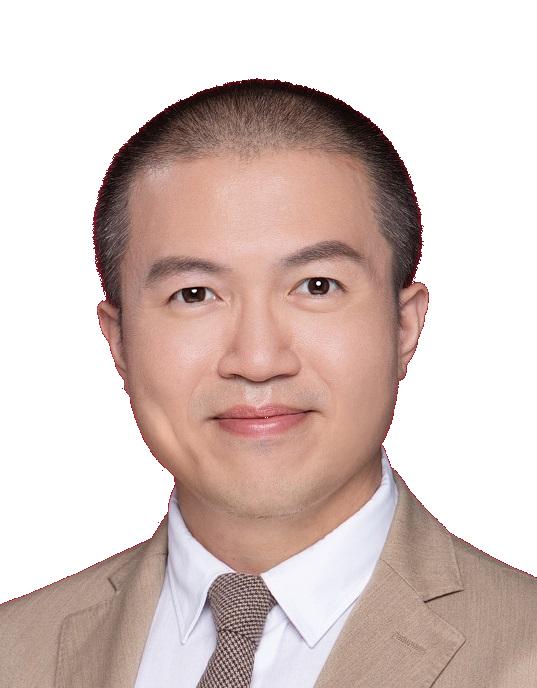}}]{Hao Li},
	associate professor and doctoral supervisor with SPEIT and the Department of Automation, Shanghai Jiao Tong University (SJTU), China. He obtained the Ph.D. degree from the Robotics Center of MINES ParisTech and INRIA in 2012, the M.Eng. degree and B.Eng. degree from the Department of Automation of SJTU in 2009 and 2006 respectively. His current research interests are in automation, computer vision, data fusion, and cooperative intelligent systems.
\end{IEEEbiography}

\end{document}